\documentclass[11pt, logo, onecolumn, copyright, colorlinks=true, allcolors=blue]{nvidiatechreport}
\usepackage{graphicx} % Required for inserting images
\usepackage{subcaption,ragged2e}
\usepackage[round]{natbib}

\usepackage[normalem]{ulem}
\usepackage[utf8]{inputenc} % allow utf-8 input
\usepackage[T1]{fontenc}    % use 8-bit T1 fonts
\usepackage{url}
\usepackage{tikz}
\usepackage{enumitem}
\usetikzlibrary{positioning}
\usepackage{array}
\usepackage{booktabs}
\usepackage{wrapfig}
\usepackage{caption} % For a better control over table captions
\usepackage{listings}
\usepackage{adjustbox}
\usepackage{footnote}
\usepackage{multirow}
\usepackage{amsmath}
\usepackage{amsfonts}
\usepackage{breakcites}
\usepackage{blindtext}
\usepackage{svg}
\usepackage[most]{tcolorbox}
\usepackage{tabularx}
\usepackage{makecell}
\usepackage{float}
\usepackage{pgf-pie}
\usepackage{multicol}
\usepackage{parskip}
\usepackage[T1]{fontenc}
\usepackage[utf8]{inputenc}
\usepackage{xspace}

\newcommand{\ourmodel}{Nemotron 3 Nano Omni\xspace}
\newcommand{\ourprevmodel}{Nemotron Nano V2 VL\xspace}
\newcommand{\ourllm}{Nemotron 3 Nano 30B-A3B\xspace}

\usepackage{tikz}
\usepackage{adjustbox} % preamble

\usetikzlibrary{positioning,arrows.meta,backgrounds}

\definecolor{stagefill}{RGB}{226,241,214}   % pale green
\definecolor{bgblue}{RGB}{154,179,211}      % light blue
\definecolor{bgblue2}{RGB}{121,159,199}     % deeper blue (bottom band)
\usetikzlibrary{positioning,arrows.meta,backgrounds,fit}
\tikzset{
  >={Stealth[length=3mm,width=2mm]},
  stage/.style={
    draw, rounded corners=2pt, fill=stagefill,
    minimum width=4.2cm, minimum height=1.9cm,
    align=center, inner sep=6pt
  },
  band/.style={
    draw=none, fill=white, rounded corners=2pt,
    minimum height=8mm, align=center, inner sep=4pt
  },
  dashedgroup/.style={
    draw=black!50, dashed, rounded corners=3pt, inner sep=8pt
  }
}

\title{\ourmodel : Efficient and Open Multimodal Intelligence}
\author{\large NVIDIA}
\date{}

\begin{document}

\begin{abstract}

\large \textbf{Abstract.} We introduce \ourmodel, the latest model in the Nemotron multimodal series and the first to natively support audio inputs alongside text, images, and video.  \ourmodel delivers consistent accuracy improvements over its predecessor, \ourprevmodel, across all modalities, enabled by advances in architecture, training data and recipes. In particular, Nemotron 3 delivers leading results in real-world document understanding, long audio-video comprehension, and agentic computer use. Built on the highly efficient \ourllm backbone, \ourmodel further incorporates innovative multimodal token-reduction techniques to deliver substantially lower inference latency and higher throughput than other models of similar size. We are releasing model checkpoints in BF16, FP8, and FP4 formats, along with portions of the training data and codebase to facilitate further research and development.
\end{abstract}

\maketitle

{\small
    {\href{https://huggingface.co/nvidia/Nemotron-3-Nano-Omni-30B-A3B-Reasoning-BF16}{Model-BF16}} \,|\,
    {\href{https://huggingface.co/nvidia/Nemotron-3-Nano-Omni-30B-A3B-Reasoning-FP8}{Model-FP8}} \,|\,
    {\href{https://huggingface.co/nvidia/Nemotron-3-Nano-Omni-30B-A3B-Reasoning-NVFP4}{Model-FP4}} \,|\,
    {\href{https://huggingface.co/datasets/nvidia/Nemotron-Image-Training-v3}{Dataset}} \,|\,
    {\href{https://github.com/NVIDIA-NeMo/Megatron-Bridge/tree/main/examples/models/vlm/nemotron_3_omni}{Megatron-Bridge}  | \href{https://github.com/NVIDIA-NeMo/RL/blob/nano-v3-omni/docs/guides/nemotron-3-nano-omni.md}{NeMo-RL}}
    \,|\,
    \href{https://github.com/NVIDIA-NeMo/DataDesigner/tree/main/docs/assets/recipes/vlm_long_doc}{Example Data Pipeline}
}

\section{Introduction}
\label{section:introduction}

In this work, we present \ourmodel, an efficient omni-modal model built on the \ourllm~\citep{nvidia2025nvidianemotron3efficient} language model backbone, augmented with the C-RADIOv4-H\footnote{\url{https://huggingface.co/nvidia/C-RADIOv4-H}}
 \citep{ranzinger2026cradiov4techreport, Heinrich_2025_CVPR} vision encoder and the Parakeet-TDT-0.6B-v2\footnote{\url{https://huggingface.co/nvidia/parakeet-tdt-0.6b-v2}}\citep{pmlr-v202-xu23g,rekesh2023fastconformerlinearlyscalable,sekoyan2025canary1bv2parakeettdt06bv3efficient} audio encoder. \ourmodel extends the Nemotron multimodal family with native audio support and improved reasoning capability across all supported modalities. It is particularly effective in practical multimodal settings, including real-world document understanding, long audio-video comprehension, and agentic computer use. In addition, \ourmodel incorporates innovative multimodal token-reduction techniques that substantially reduce inference latency and increase throughput, enabling efficient deployment without sacrificing model quality.

Compared to our previous release, \ourprevmodel~\citep{nvidia2025nvidianemotronnanov2}, \ourmodel introduces several key design choices and architectural advances:
\begin{enumerate}
\item \textbf{Improved LLM Backbone.} We replace the dense Nemotron Nano V2 12B hybrid backbone with the \ourllm Mixture-of-Experts (MoE) hybrid backbone, enabling more efficient processing of long multimodal sequences and higher inference throughput.
\item \textbf{Native Audio Support.} We extend the model to natively support audio inputs in addition to text, images, and video.
\item \textbf{Dynamic Image Resolution.} We replace the tiling-based image processing approach with a dynamic resolution strategy that better preserves native aspect ratios.
\item \textbf{Temporal Video Compression.} We introduce Conv3D-based temporal compression for video, achieving a 2$\times$ reduction in temporal tokens.
\item \textbf{Extended Context Length.} We increase the maximum context length from 128K to 256K tokens, improving performance on long-context multimodal reasoning tasks.
\end{enumerate}

Training an omni-modal MoE model introduces challenges in modality alignment, training stability, and data balancing across heterogeneous sources. To preserve the strong text reasoning capabilities of the base LLM while improving multimodal performacne, we adopt a multi-stage training strategy that progressively introduces new modalities and scales context length. This staged approach mitigates catastrophic forgetting and stabilizes cross-modal alignment during training.

Driven by these technical improvements, \ourmodel achieves substantial gains over \ourprevmodel across a wide range of tasks. In particular, it attains leading results in document understanding, audio-visual reasoning, and audio benchmarks, ranking at or near the top of leaderboards such as OCRBench-V2 \citep{Liu_2024}, MMLongBench-DOC \citep{ma2024mmlongbenchdocbenchmarkinglongcontextdocument}, VoiceBench~\citep{voicebench},  WorldSense~\citep{worldsense}, and DailyOmni~\citep{zhou2025dailyomni}.

These improvements also translate into higher inference efficiency and lower latency. On NVIDIA B200,
\ourmodel achieves 3$\times$ higher single-stream output token throughput than Qwen3-Omni \citep{xu2025qwen3omnitechnicalreport} and
9$\times$ higher output token throughput per GPU at a fixed interactivity target. Compared with
\ourprevmodel, \ourmodel provides 3$\times$ higher throughput at the same interactivity target and
2$\times$ higher single-stream output token throughput. \ourmodel ranks as the most cost-efficient open video understanding model on \href{https://mediaperf.org/leaderboard}{MediaPerf}.

Along with this report, we are releasing the model checkpoints on HuggingFace:
\begin{itemize}
    \item \href{https://huggingface.co/nvidia/Nemotron-3-Nano-Omni-30B-A3B-Reasoning-BF16}{Nemotron-3-Nano-Omni-30B-A3B-Reasoning-BF16}
    \item \href{https://huggingface.co/nvidia/Nemotron-3-Nano-Omni-30B-A3B-Reasoning-FP8}{Nemotron-3-Nano-Omni-30B-A3B-Reasoning-FP8}
    \item \href{https://huggingface.co/nvidia/Nemotron-3-Nano-Omni-30B-A3B-Reasoning-NVFP4}{Nemotron-3-Nano-Omni-30B-A3B-Reasoning-NVFP4}
\end{itemize}

We are also releasing part of our training datasets, pipelines, and code:
\begin{itemize}
    \item \href{https://huggingface.co/datasets/nvidia/Nemotron-Image-Training-v3}{Nemotron-Image-Training-v3}: A collection of $\sim$6.9 million training samples.
    \item \href{https://github.com/NVIDIA-NeMo/DataDesigner/tree/main/docs/assets/recipes/vlm_long_doc}{Examples of data generation pipelines}
    \item \href{https://github.com/NVIDIA-NeMo/Megatron-Bridge/tree/main/examples/models/vlm/nemotron_3_omni}{Training code on Megatron-Bridge}
    \item \href{https://github.com/NVIDIA-NeMo/RL/blob/nano-v3-omni/docs/guides/nemotron-3-nano-omni.md}{NeMo RL guide for \ourmodel}
\end{itemize}

The remainder of this paper is organized as follows. Section~\ref{section:model_arch} describes the model architecture. Section~\ref{section:recipe} details the training pipeline and datasets. Section~\ref{section:experiments} presents evaluation results across all modalities.
\section{Model Architecture}
\label{section:model_arch}

\begin{figure}[htpb]
  \centering
  \includegraphics[width=0.7\linewidth]{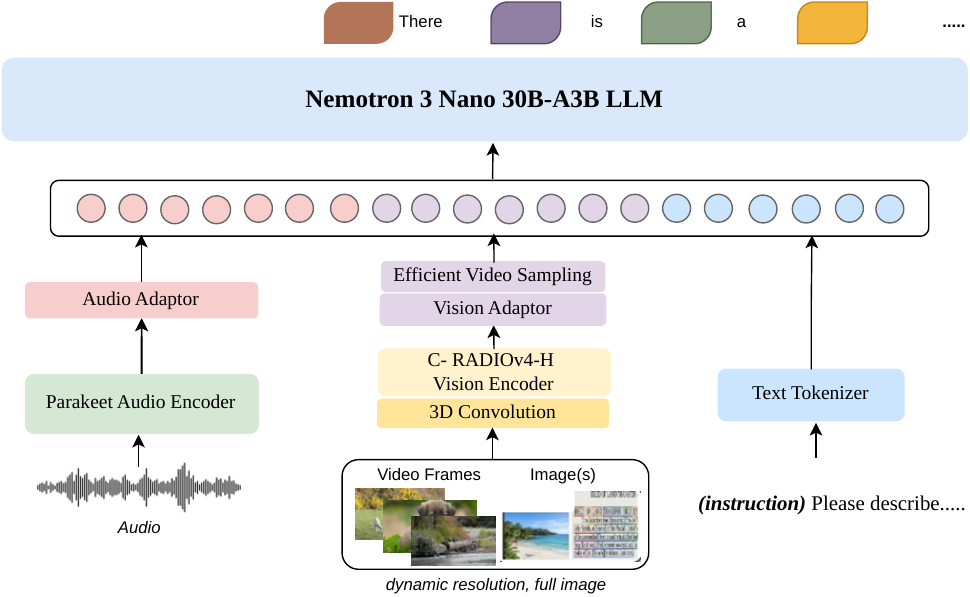} % or omit extension: {figures/VLMs}
  \caption{\ourmodel architecture. For encoding images and videos we use dynamic resolution. Additionally, videos use Conv3D and optionally Efficient Video Sampling for higher throughput. Audio inputs are encoded using Parakeet v2 audio encoder. Visual, audio, and text tokens are concatenated and fed to the LLM.}
  \label{fig:architecture}
\end{figure}

Our model follows an encoder-projector-decoder design, combining the \ourllm \citep{nvidia2025nvidianemotron3efficient} language model with modality-specific encoders for vision and audio, connected via MLP projectors. An overview of the architecture is shown in Figure~\ref{fig:architecture}. The vision encoder is based on C-RADIOv4-H \citep{ranzinger2026cradiov4techreport, Heinrich_2025_CVPR}, while the audio encoder is initialized with Parakeet-TDT-0.6B-v2 \citep{pmlr-v202-xu23g,rekesh2023fastconformerlinearlyscalable,sekoyan2025canary1bv2parakeettdt06bv3efficient}.

To handle varying image resolutions, we replace the tiling strategy used in \ourprevmodel\citep{nvidia2025nvidianemotronnanov2} with dynamic resolution processing that preserves the native aspect ratio. Each image is decomposed into a variable number of $16 \times 16$ patches, with the total number of visual tokens per image constrained between 1,024 and 13,312. This equates to an image size of $512 \times 512$ and $1840 \times 1840$, respectively, for square images. Prior to projection, we apply a pixel shuffle operation with 4$\times$ downsampling to reduce the token count presented to the language model. For video frames, we use a Conv3D patch embedder that compresses every two frames into one, leading to a $2 \times$ reduction in the total number of tokens for video inputs.

Audio inputs are resampled to 16 kHz mono and encoded using the Parakeet-TDT-0.6B-v2 FastConformer encoder. We first compute log-mel spectrogram features with a 10 ms hop size, followed by three stride-2 convolutional subsampling layers, resulting in an overall $\sim$8$\times$ temporal downsampling. This yields approximately 12.5 tokens per second of audio (i.e., $\sim$80 ms per token). Audio streams are segmented into 30-second clips (corresponding to $\sim$375 tokens per clip) with the last clip accounting for the remainder. Streams shorter than 30 seconds are not padded. We train the model to handle inputs ranging from 0.5 second to 20 minutes, but the model context length can accommodate audio input of over 5 hours.

For multimodal inputs containing both visual and audio streams (e.g., videos with audio), modality tokens are interleaved in temporal order during sequence construction to enable joint temporal reasoning across modalities.
\section{Training Recipe \& Datasets}
\label{section:recipe}

% Preamble
\usetikzlibrary{arrows.meta,fit,backgrounds,calc}

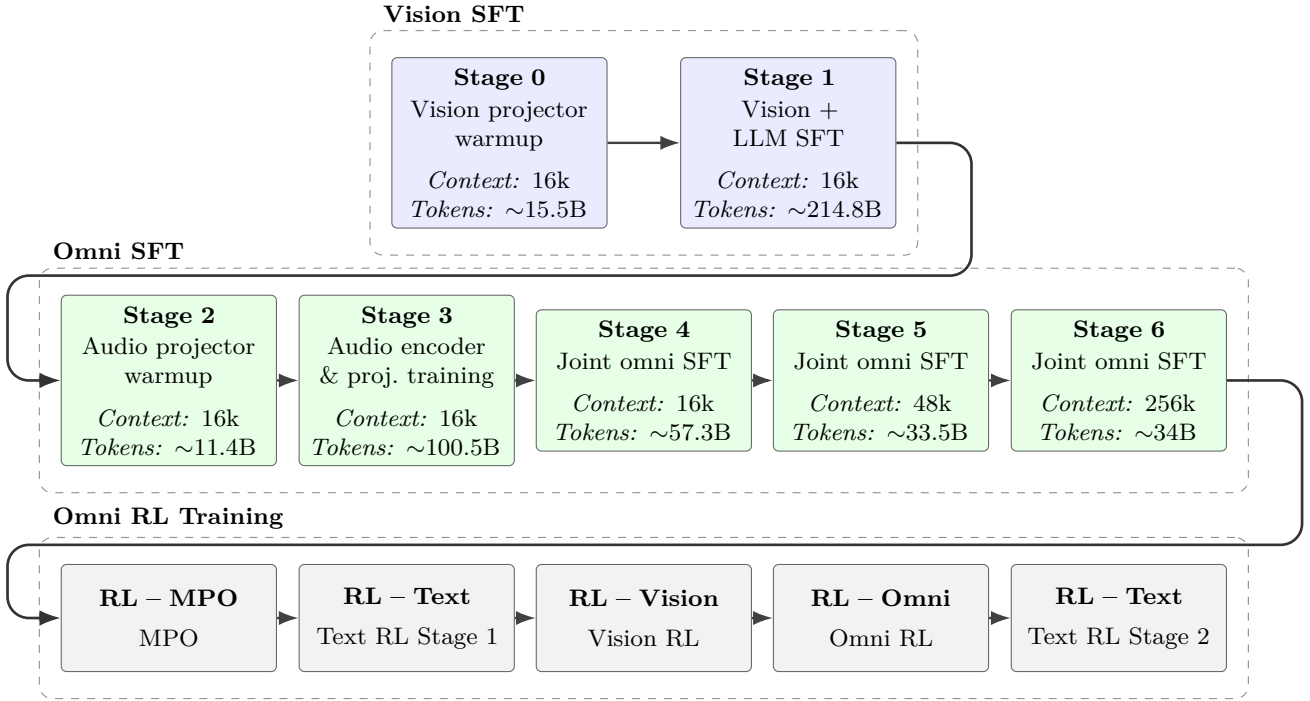
\begin{figure*}[t]
\centering
\begin{tikzpicture}[
  >=Latex,
  stage/.style={
    draw=black!60,
    rounded corners=2pt,
    align=center,
    font=\footnotesize,
    text width=2.55cm,
    minimum height=1.42cm,
    inner sep=4pt
  },
  vision/.style={stage, fill=blue!8},
  omni/.style={stage, fill=green!10},
  rl/.style={stage, fill=gray!10},
  flow/.style={
    -{Latex[length=2.6mm,width=1.9mm]},
    line width=0.9pt,
    draw=black!75
  },
  transfer/.style={
    -{Latex[length=2.8mm,width=2.1mm]},
    line width=1.0pt,
    draw=black!80,
    rounded corners=8pt
  },
  groupbox/.style={
    draw=black!45,
    dashed,
    rounded corners=5pt,
    inner xsep=8pt,
    inner ysep=10pt   % extra padding so transfer arrows can live here
  },
  grouplabel/.style={
    font=\footnotesize\bfseries,
    fill=white,
    inner xsep=3pt,
    inner ysep=1pt
  }
]

% ---------------- Top row: Vision SFT ----------------
\node[vision] (s0) at (-1.90, 0.00) {
  \textbf{Stage 0}\\[1pt]
  Vision projector\\
  warmup\\[5pt]
  \textit{Context:} 16k\\
  \textit{Tokens:} $\sim$15.5B
};

\node[vision] (s1) at (1.90, 0.00) {
  \textbf{Stage 1}\\[1pt]
  Vision + \\
  LLM SFT\\[5pt]
  \textit{Context:} 16k\\
  \textit{Tokens:} $\sim$214.8B
};

% ---------------- Middle row: Omni SFT ----------------
\node[omni] (s2) at (-6.25, -3.15) {
  \textbf{Stage 2}\\[1pt]
  Audio projector\\
  warmup\\[5pt]
  \textit{Context:} 16k\\
  \textit{Tokens:} $\sim$11.4B
};

\node[omni] (s3) at (-3.12, -3.15) {
  \textbf{Stage 3}\\[1pt]
  Audio encoder\\
  \& proj.\ training\\[5pt]
  \textit{Context:} 16k\\
  \textit{Tokens:} $\sim$100.5B
};

\node[omni] (s4) at (0.00, -3.15) {
  \textbf{Stage 4}\\[1pt]
  Joint omni SFT\\[5pt]
  \textit{Context:} 16k\\
  \textit{Tokens:} $\sim$57.3B
};

\node[omni] (s5) at (3.12, -3.15) {
  \textbf{Stage 5}\\[1pt]
  Joint omni SFT\\[5pt]
  \textit{Context:} 48k\\
  \textit{Tokens:} $\sim$33.5B
};

\node[omni] (s6) at (6.25, -3.15) {
  \textbf{Stage 6}\\[1pt]
  Joint omni SFT\\[5pt]
  \textit{Context:} 256k\\
  \textit{Tokens:} $\sim$34B
};

% ---------------- Bottom row: Omni RL Training ----------------
\node[rl] (mpo) at (-6.25, -6.30) {
  \textbf{RL -- MPO}\\[4pt]
  MPO
};

\node[rl] (rltxt1) at (-3.12, -6.30) {
  \textbf{RL -- Text}\\[4pt]
  Text RL Stage 1
};

\node[rl] (rlimg) at (0.00, -6.30) {
  \textbf{RL -- Vision}\\[4pt]
  Vision RL
};

\node[rl] (rlomni) at (3.12, -6.30) {
  \textbf{RL -- Omni}\\[4pt]
  Omni RL
};

\node[rl] (rltxt2) at (6.25, -6.30) {
  \textbf{RL -- Text}\\[4pt]
  Text RL Stage 2
};

% ---------------- Group boxes ----------------
\begin{scope}[on background layer]
  \node[groupbox, fit=(s0)(s1)] (gVision) {};
  \node[groupbox, fit=(s2)(s3)(s4)(s5)(s6)] (gOmni) {};
  \node[groupbox, fit=(mpo)(rltxt1)(rlimg)(rlomni)(rltxt2)] (gRL) {};
\end{scope}

% ---------------- Group labels ----------------
\node[grouplabel, anchor=south west]
  at ([xshift=2pt,yshift=2pt]gVision.north west) {Vision SFT};

\node[grouplabel, anchor=south west]
  at ([xshift=2pt,yshift=2pt]gOmni.north west) {Omni SFT};

\node[grouplabel, anchor=south west]
  at ([xshift=2pt,yshift=2pt]gRL.north west) {Omni RL Training};

% ---------------- In-row arrows ----------------
\draw[flow] (s0.east) -- (s1.west);

\draw[flow] (s2.east) -- (s3.west);
\draw[flow] (s3.east) -- (s4.west);
\draw[flow] (s4.east) -- (s5.west);
\draw[flow] (s5.east) -- (s6.west);

\draw[flow] (mpo.east) -- (rltxt1.west);
\draw[flow] (rltxt1.east) -- (rlimg.west);
\draw[flow] (rlimg.east) -- (rlomni.west);
\draw[flow] (rlomni.east) -- (rltxt2.west);

% ---------------- Transfer lanes ----------------
% Put the transfer arrows in the empty top padding of the next row's dashed box.
\coordinate (omniLane) at ([yshift=-3pt]gOmni.north);
\coordinate (rlLane)   at ([yshift=-3pt]gRL.north);

\coordinate (visionOut) at ([xshift=7mm]gVision.east);
\coordinate (omniOut)   at ([xshift=7mm]gOmni.east);

\coordinate (s2Entry)   at ([xshift=-7mm]s2.west);
\coordinate (mpoEntry)  at ([xshift=-7mm]mpo.west);

% Vision SFT -> Omni SFT
\draw[transfer]
  (s1.east) -- (visionOut)
  -- (visionOut |- omniLane)
  -- (s2Entry |- omniLane)
  -- (s2Entry)
  -- (s2.west);

% Omni SFT -> Omni RL Training
\draw[transfer]
  (s6.east) -- (omniOut)
  -- (omniOut |- rlLane)
  -- (mpoEntry |- rlLane)
  -- (mpoEntry)
  -- (mpo.west);

\end{tikzpicture}
\caption{
Staged training recipe for the v3 omni-modal model. The pipeline first performs vision SFT, then joint omni SFT while progressively extending context length, followed by omni-modal RL training.
}
\label{fig:v3-omni-stage-flow}
\end{figure*}

Training an omni-modal reasoning model with heterogeneous encoders requires careful orchestration. To this end, we adopt a staged training strategy that first performs supervised fine-tuning (SFT) to progressively align modalities, improve multi-modal instruction-following and extend context capacity, followed by reinforcement learning (RL) to further refine reasoning and safety. Figure~\ref{fig:v3-omni-stage-flow} illustrates the overall progression of these stages.

\subsection{SFT}

Our SFT pipeline is split into seven stages that progressively introduce new modalities and increase context length. This curriculum is designed to promote stable cross-modal alignment and mitigate catastrophic forgetting while improving multi-modal understanding. Detailed descriptions of each stage are provided in Sections~\ref{sec:sft-stage0}–\ref{sec:sft-stage6}, and an overview is provided in Table~\ref{tab:sft_data_summary}.
\begin{table}[t]
\centering
\resizebox{\textwidth}{!}{%
\begin{tabular}{lcccl}
\toprule
& \textbf{Number of Samples} & \textbf{Number of Tokens} & \textbf{Primary Data Domains} \\
\midrule
Stage 0 & 9.35M & 15.5B  & Captioning, OCR, document, VQA \\
Stage 1 & 86.3M & 214.8B & Comprehensive vision-language SFT \\
Stage 2 & 59.2M & 11.4B  & ASR (Granary) \\
Stage 3 & 242.0M & 100.5B & ASR, sound, music, speech understanding \\
Stage 4 & 30.5M & 57.3B  & Vision, video, audio, text, omni, safety \\
Stage 5 & 6.08M & 33.5B  & Long video, omni, reasoning \\
Stage 6 & 623K  & 34.0B  & Ultra-long documents, long-context text \\
\midrule
\textbf{Total (all stages)} & \textbf{434.1M} & \textbf{466.9B} & & \\
\bottomrule
\end{tabular}%
}
\caption{Approximate values for the total number of samples and tokens (including masked tokens from the prompt) in the training datasets across the SFT stages. This includes any sample repetitions.}
\label{tab:sft_data_summary}
\end{table}

\subsubsection{Stage 0: Vision projector warmup}
\label{sec:sft-stage0}

We begin by training only the vision MLP projector to align the vision and language modalities with a maximum context length of 16384, while keeping all other components frozen. This stage uses approximately 9.35 million vision–text samples ($\sim$ 15.5B tokens), including a portion of the Stage 1 dataset (see Section~\ref{sec:sft-stage1}) and covering a diverse set of tasks, including image captioning, visual grounding, OCR, document understanding, GUI understanding, and general visual question answering.

\subsubsection{Stage 1: Vision SFT 16k}
\label{sec:sft-stage1}

After training the vision projector, we unfreeze both the language model and the vision encoder for joint vision–language fine-tuning. During this stage, the model develops its core vision-language capabilities. The training data builds upon the SFT Stage 1 dataset used in \ourprevmodel~\citep{nvidia2025nvidianemotronnanov2}, with several key enhancements.

First, we replace its text-only subset with a portion of the SFT dataset from \ourllm, resulting in higher-quality text reasoning samples. Second, we improve label quality by re-annotating noisy subsets using models from the Qwen3-VL series \citep{yang2025qwen3technicalreport}. Third, we enhance the availability and quality of reasoning traces by incorporating both human-annotated and model-generated chains of thought, leveraging models from the Qwen3-VL \citep{yang2025qwen3technicalreport}, Qwen3.5 \citep{qwen3.5}, and Kimi-K2.5 \citep{kimiteam2026kimik25visualagentic} families.

Finally, we expand coverage across domains, including GUI understanding, visual grounding, charts, tables, document understanding, and video understanding, as well as across multiple languages. This is achieved through a combination of publicly available datasets, as well as internally curated data, including human annotation. To increase domain coverage, we additionally develop fully-synthetic data pipelines ensuring broad representation across domains, question types, and visual diversity. Guided by the gaps identified in the training blend, we source relevant data and generate synthetic question-answer pairs at scale using frontier open-source models such as Qwen3-VL \citep{yang2025qwen3technicalreport}, Qwen 3.5 \citep{qwen3.5}, GPT-OSS \citep{gpt-oss}, Nemotron-Parse \citep{chumachenko2025nvidia}, and DeepSeek-OCR \citep{wei2025deepseek}. For each domain, we generate question-answer pairs from images, videos, or OCR extracted from images using domain-specific instructions. This is followed by distillation of reasoning traces and strict filtering of the resulting samples to ensure data correctness, usefulness, and overall quality.

The resulting dataset comprises approximately 86.3M samples ($\sim$ 214.8B tokens), including sample repetitions.

\subsubsection{Stage 2: Audio projector warmup}
\label{sec:sft-stage2}

Analogous to Stage~0 for vision, this stage warms up the audio MLP projector~\citep{chen2024salm} while keeping the LLM, vision encoder, and Parakeet-TDT audio encoder all frozen.

The training data consists of the Granary v1.1 ASR dataset~\citep{koluguri2025granary}, comprising approximately 59.2M samples ($\sim$11.4B tokens) of diverse automatic speech recognition data across varied acoustic conditions, speaking styles, and languages.

\subsubsection{Stage 3: Audio projector \& encoder}
\label{sec:sft-stage3}

Building on Stage~2, this stage unfreezes the Parakeet-TDT audio encoder while keeping the LLM backbone and vision encoder frozen. The audio encoder and its associated projector are jointly trained on an expanded audio corpus.

As shown in Table \ref{tab:stage3_data}, this stage is trained using a mixture of ASR data along with sound, music, and speech understanding. Audio samples are paired with captions, multiple-choice questions, and open-ended questions, with a subset further augmented with reasoning traces. Our synthetic data generation pipeline leverages open models like Qwen3-Omni-30B-A3B to produce captions and specialized music tools to produce metadata. These outputs are then used to generate QA pairs via GPT-OSS-120B.

\begin{table}[t]
\centering
\resizebox{0.8\linewidth}{!}{%
\begin{tabular}{lrrr}
\toprule
\textbf{Dataset type} & \textbf{Number of samples} & \textbf{\% of total tokens} & \textbf{Number of tokens} \\
\midrule
ASR                   & 113.8M & 22.7\% & 22.8B \\
Sound understanding   & 61.0M  & 24.4\% & 24.5B \\
Music understanding   & 19.8M  & 43.3\% & 43.5B \\
Speech understanding  & 47.5M  & 9.6\%  & 9.6B \\
\midrule
\textbf{Total}        & \textbf{242.0M} &  & \textbf{100.5B} \\
\bottomrule
\end{tabular}%
}
\caption{Dataset composition for the audio pretraining stage.}
\label{tab:stage3_data}
\end{table}

%Training uses a learning rate of $2.5{\times}10^{-5}$ with cosine decay to 0, weight decay of 0.05, a global batch size of 512, and scales to 128 nodes (1,024 H100 GPUs). The decoder sequence length is 16,384 tokens.

\subsubsection{Stage 4: Omni SFT 16k}
\label{sec:sft-stage4}

This is the first stage that jointly trains on all modalities. All model parameters, including the LLM backbone, are trainable. The data mixture combines vision SFT, text instruction following, safety, video understanding, omni (audio+video) QA and captioning, ASR, and audio reasoning data.

\begin{table}[t]
\centering
\resizebox{0.8\linewidth}{!}{%
\begin{tabular}{lrrr}
\toprule
\textbf{Dataset type} & \textbf{Number of samples} & \textbf{\% of total tokens} & \textbf{Number of tokens} \\
\midrule
Vision data                  & 14.6M & 53.4\%  & 30.6B \\
Text data                    & 948K  & 6.1\%   & 3.5B \\
Text safety data             & 14K   & 0.02\%  & 10.4M \\
Image safety data            & 9K    & 0.02\%  & 10.0M \\
Short video data             & 1.3M  & 11.0\%  & 6.3B \\
Short video reasoning data   & 388K  & 4.2\%   & 2.4B \\
Short video omni data        & 251K  & 2.8\%   & 1.6B \\
ASR data                     & 2.9M  & 1.1\%   & 640M \\
Audio reasoning data         & 765K  & 4.4\%   & 2.5B \\
Audio data                   & 9.3M  & 16.9\%  & 9.7B \\
\midrule
\textbf{Total}               & \textbf{30.5M} &  & \textbf{57.3B} \\
\bottomrule
\end{tabular}%
}
\caption{Dataset composition for Stage~4: Joint Omni SFT at 16k context length.}
\label{tab:stage4_data}
\end{table}

% \begin{table}[t]
% \centering
% \caption{Stage~4 (Omni SFT 16k) data composition. Effective samples and token estimates account for repetition factors.}
% \label{tab:stage4_data}
% \resizebox{\textwidth}{!}{%
% \begin{tabular}{llrrr}
% \toprule
% \textbf{Category} & \textbf{Dataset} & \textbf{Eff.\ Samples} & \textbf{Est.\ Tokens} \\
% \midrule
% Vision  & Eagle SFT v13.77                    & 72.6M & 152.3B \\
% Text    & Nickel Capybara 16k                  & 4.7M  & 15.9B \\
% Safety  & Text safety + image safety            & 116K  & 109M \\
% Vision  & SFT-0.5 (BenchFit QA, dense OCR QA)  & 542K  & 458M \\
% Video   & Short video QA                        & 2.7M  & 9.2B \\
% Video   & Video QA v1+v2                        & 1.1M  & 4.4B \\
% Video   & Video reasoning (thinking traces)     & 1.6M  & 6.0B \\
% Video   & Dense captions (short, no think)       & 2.6M  & 17.1B \\
% Omni    & Short omni QA + captions              & 1.3M  & 7.9B \\
% ASR     & Granary v1.1 (small + full subsets)    & 14.5M & 3.0B \\
% Audio   & Audio reasoning SFT + BBH reasoning   & 3.8M  & 12.1B \\
% Audio   & Audio SFT                             & 45.9M & 45.8B \\
% Audio   & Audio reasoning (MMAU)                & 1.5M  & 609M \\
% \midrule
% \multicolumn{2}{l}{\textbf{Total}} & \textbf{153.3M} & \textbf{$\sim$281B} \\
% \bottomrule
% \end{tabular}%
% }
% \end{table}

Table~\ref{tab:stage4_data} summarizes the data composition. The dominant sources are the vision dataset (30.6B tokens), the audio dataset (9.7B tokens), and the short video data (6.3B tokens). The omni-modal data used in this stage is a blend of audio-visual captions, open-ended QA and MCQ style QA. Videos less than 2 minutes length are used as source media for this data. The question-answer pairs and captions are synthetically generated by first extracting audio-visual metadata from videos and then using that metadata for question-answer generation and summarization using open-source models Qwen3-Omni-30B-A3B  and GPT-OSS-120B. The audio reasoning dataset comprises speech-to-text conversations synthesized by converting text SFT user turns into spoken form and generating LLM responses to a curated subset of ASR prompts.

% Video inputs use up to 64 frames with conv3d temporal compression and video augmentation that scales frame count up by $4{\times}$ while inversely scaling spatial resolution.

% Training uses a learning rate of $10^{-5}$ with cosine decay to $10^{-7}$, a global batch size of 512, and runs on 64 nodes (512 H100 GPUs) with a decoder sequence length of 16,384 tokens. The MoE auxiliary loss coefficient is $10^{-9}$.

\subsubsection{Stage 5: Omni SFT 48k}
\label{sec:sft-stage5}
% \ehsan{will do}

This stage extends the context length to 49,152 tokens with all model parameters trainable. The data mixture is rebalanced to emphasize longer sequences, with reduced sampling of short-context data and increased weight on medium and long video, omni, and reasoning data.

\begin{table}[t]
\centering
\resizebox{0.8\linewidth}{!}{%
\begin{tabular}{lrrr}
\toprule
\textbf{Category} & \textbf{Number of samples} & \textbf{\% of total tokens} & \textbf{Number of Tokens} \\
\midrule
ASR                 & 650K   & 0.4\%  & 0.12B \\
Audio               & 2.84M  & 11.3\% & 3.80B \\
Vision              & 1.17M  & 9.8\%  & 3.28B \\
Text                & 101K   & 7.2\%  & 2.42B \\
Safety              & 45K    & 0.1\%  & 0.04B \\
Video (short)       & 25K    & 0.6\%  & 0.21B \\
Video (medium)      & 96K    & 5.8\%  & 1.95B \\
Video (long)        & 74K    & 3.3\%  & 1.11B \\
Video reasoning     & 167K   & 10.2\% & 3.42B \\
Omni (short)        & 6K     & 0.3\%  & 0.09B \\
Omni (medium+long)  & 710K   & 39.1\% & 13.10B \\
Omni reasoning      & 198K   & 11.8\% & 3.94B \\
\midrule
\textbf{Total}      & \textbf{6.08M} & \textbf{100\%} & \textbf{$\sim$33.5B} \\
\bottomrule
\end{tabular}%
}
\caption{Stage~5 (Omni SFT 48k) data composition by category.}
\label{tab:stage5_data}
\end{table}

Table~\ref{tab:stage5_data} shows the per-category breakdown. Compared to Stage~4, this stage has a much higher proportion of long-context data: medium and long video, omni data with joint audio-video understanding, and reasoning traces. Short video and omni data are downsampled, while medium/long omni data and reasoning data receive the bulk of the training budget. To improve safety coverage, we additionally incorporate the safety data blend from Nemotron 3 Super \citep{nvidia2026nemotron3superopen}.

For the 48k SFT stage, omni-modal data comprising reasoning and non-reasoning single-turn QA is synthesized from diverse domains and categories. The pipeline segments videos into 20-second clips, extracts audio-visual metadata using multimodal models such as Qwen3-Omni-30B-A3B, and generates QA pairs and reasoning traces via open-source reasoning models such as GPT-OSS-120B.

\subsubsection{Stage 6: Omni SFT 256k}
\label{sec:sft-stage6}

This stage extends the context length to 262,144 and is intended to significantly increase the model's long context capabilities. The data for this stage consists of $\sim$34.0B tokens across long-context text-only and vision domains such as long-context reasoning and long document understanding (see Table \ref{tab:stage6_data}). It particularly improves the model’s ability to analyze documents spanning 10 to 100+ pages, including reasoning over text, charts, and complex tables. We assemble a diverse collection of long-form documents, including academic papers, financial reports, and presentations, and leverage vision-language models to generate synthetic question-answer pairs and reasoning traces at the page, multi-page, and full-document levels. To support long-document understanding, we release runnable data pipeline recipes\footnote{\url{https://github.com/NVIDIA-NeMo/DataDesigner/tree/main/docs/assets/recipes/vlm_long_doc}} using NeMo Data Designer \citep{nemo-data-designer}.

\begin{table}[t]
\centering
\resizebox{0.8\linewidth}{!}{%
\begin{tabular}{lrrr}
\toprule
\textbf{Dataset type} & \textbf{Number of samples} & \textbf{\% of total tokens} & \textbf{Number of tokens} \\
\midrule
Long Context Vision & 508K & 90.9\% & 30.9B \\
Text                & 63K  & 7.3\%  & 2.5B \\
Long Context Text   & 2.2K & 1.5\%  & 506M \\
Vision              & 50K  & 0.3\%  & 106M \\
\midrule
\textbf{Total}      & \textbf{623K} & \textbf{100\%} & \textbf{34.0B} \\
\bottomrule
\end{tabular}%
}
\caption{Dataset composition for the ultra-long context Stage 6.}
\label{tab:stage6_data}
\end{table}

The audio encoder and projector are frozen during this stage to focus model capacity on long-context text and document understanding.

\subsubsection{Training Details}

\begin{table}[t!]\small
\centering
\begin{tabular}{@{}l|c|c|c|c|c|c|c@{}}
\toprule
& \textbf{Stage 0} & \textbf{Stage 1} & \textbf{Stage 2} & \textbf{Stage 3} & \textbf{Stage 4} & \textbf{Stage 5} & \textbf{Stage 6} \\ \toprule
\texttt{Context Length}  & \multicolumn{5}{c|}{16K} & 48k & 256k  \\ \hline
\texttt{Max Video Frames} & -- & 64 & -- & -- & 64 & 256 & 256 \\ \hline
\texttt{Global BS}  & 128  & 256 & \multicolumn{3}{c|}{512} & 256 & 128 \\ \hline
% \texttt{TP}  & \multicolumn{7}{c}{2} \\ \hline
% \texttt{EP}  & \multicolumn{7}{c}{32} \\ \hline
\texttt{CP}  & \multicolumn{5}{c|}{-} & 2 & 16 \\ \hline
\texttt{LR} & $10^{-3}$ & $5\times10^{-5}$ & $10^{-3}$ & $2.5\times10^{-5}$ & $10^{-5}$ & \multicolumn{2}{c}{$10^{-6}$} \\ \hline
\texttt{Minimum LR} & $10^{-5}$ & 0 & $10^{-5}$ & 0 & $10^{-7}$ & \multicolumn{2}{c}{0} \\ \hline
\makecell[l]{\texttt{Linear Warmup}\\\texttt{Fraction}} & \multicolumn{5}{c|}{0.1} & 0.01 & 0.1 \\ \hline
\texttt{Weight Decay} & 0.01 & 0.05 & 0.01 & \multicolumn{4}{c}{0.05} \\ \hline
\makecell[l]{\texttt{Trainable}\\\texttt{Modules}} & \makecell{Vision \\ Projector} & \makecell{All\\except\\audio} & \makecell{Audio \\ Projectior}  & \makecell{Audio\\Encoder \&\\Projector} & \multicolumn{2}{c|}{All} & \makecell{All\\except\\audio} \\ \hline
\texttt{\# GPU Nodes} & 32 & \multicolumn{2}{c|}{64} & 128 & \multicolumn{3}{c}{64} \\
\bottomrule
\end{tabular}
\caption{Summary of the SFT training hyperparameters. All stages use the AdamW optimizer ($\beta_1{=}0.9$, $\beta_2{=}0.999$), a cosine LR decay, BF16 precision, TP=2 and EP=32}
\label{tab:sft_hyperparams}
\end{table}

All SFT stages are trained using the Megatron framework \citep{megatron-lm}\footnote{\url{https://github.com/NVIDIA/Megatron-LM}}
, together with Transformer Engine\footnote{\url{https://github.com/NVIDIA/TransformerEngine}}
 and the Megatron Energon\footnote{\url{https://github.com/NVIDIA/Megatron-Energon}}
 dataloader. Training is conducted on 32--128 nodes of NVIDIA H100 GPUs, depending on the stage.
 
 We employ 2-way tensor parallelism (TP), 32-way expert parallelism (EP), and sequence parallelism to efficiently scale training. All stages are trained in BF16 mixed precision and use online sequence packing with a balanced greedy knapsack algorithm to maximize GPU utilization.

To fit long sequences in GPU memory, we use selective activation recomputation for the LLM backbone (recomputing core attention, MLP, LayerNorm, and MoE activations) and full block-level recomputation for all 32 vision encoder layers. Sound model activations are recomputed starting from Stage~4. Vision projection and sound projection recomputation are enabled from Stage~5 onward to support the increased memory requirements of longer sequences. Additionally, context parallelism is introduced in later stages, with 2-way and 16-way CP in Stages~5 and~6, respectively, to accommodate increasingly long sequence lengths.

The vision encoder's CPE layers are kept in eval mode in stages 1, 4 and 5 to stabilize training. For videos, we sample up to 64 frames in Stages 1 and 4, and up to 256 frames in Stages 5 and 6. We also employ video augmentation that randomly selects the target number of patches per video frame from \(\{256, 512, 768, 1024\}\). This allows us to reduce the image resolution at inference time, while scaling up the number of frames, to improve temporal information without increasing the number of tokens. We use the AdamW optimizer with $\beta_1$ and $\beta_2$ set to $0.9$ and $0.999$, respectively, and a cosine annealing schedule with a linear warmup. Table~\ref{tab:sft_hyperparams} summarizes the training hyperparameters for SFT stages.

\subsection{Reinforcement Learning}

After SFT, we apply multiple rounds of reinforcement learning to further improve instruction following, reasoning, and safety-alignment for text, image, and video modalities.
We design a curriculum learning pipeline for post-training: (1) Preference Optimization, (2) Text-RL-stage-1, (3) Image-RL, (4) Omni-RL, and (5) Text-RL-stage-2.

% Following SFT, we apply four rounds of reinforcement learning to further improve model performance. Our approach combines mixed preference optimization (MPO) with modality-specific RL techniques to enhance reasoning, instruction following, and safety across modalities.

\subsubsection{Preference Optimization}
To align our model using both preference-level and quality-level supervision, we adopt Mixed Preference Optimization (MPO) \citep{wang2024enhancingreasoningabilitymultimodal}, which combines a preference loss and a quality loss during the offline reinforcement learning stage. Specifically, we employ Direct Preference Optimization (DPO) ~\citep{rafailov2023directpreferenceoptimizationlanguage} as the preference loss and Binary Classifier Optimization (BCO) \citep{wang2024enhancingreasoningabilitymultimodal} as the quality loss. To construct the training data, we apply rejection sampling to generate candidate responses in the vision domain and assign binary labels based on outcome correctness, yielding positive samples for accepted responses and negative samples for rejected ones.

\subsubsection{Text-RL}

During text-only RL, we only train the LM parameters of the model via multi-environment RLVR/RLHF for improving general capabilities.
We reuse the RL data and infrastructure from the post-training of Nemotron 3 Nano and Super \citep{nvidia2025nemotron3nanoopen, nvidia2026nemotron3superopen}.
As part of our staged multi-modal training, during text-only RL stages we additionally freeze the LM input token embedding parameters to mitigate representational drift between multi-modal stages.

\subsubsection{Image RL}
ImageRL is the first stage of our multimodal RL pipeline. We employ outcome-based RL on visual reasoning tasks, which can be divided into the following categories

\begin{itemize}
    \item \emph{Chart, document, and text-rich image reasoning}: numerical, comparative, and trend reasoning over plots, tables, diagrams, infographics, and natural images containing text ($\sim$28K).
    \item \emph{STEM and mathematical problems}: geometry, algebra, functions, and counting, in both English and Chinese ($\sim$19K).
    \item \emph{Game and puzzle reasoning}: rule-based reasoning over rendered game-board states ($\sim$12K).
    \item \emph{Visual question answering}: open-ended and multiple-choice questions covering spatial relations, attribute recognition, and yes/no judgements ($\sim$8K).
    \item \emph{Visual grounding}: click-coordinate prediction on desktop, mobile, and web screenshots ($\sim$7K).
\end{itemize}

During training, each prompt is graded by a $[0,1]$ scalar that linearly combines an outcome score.
The outcome score comes from one of four rule-based verifiers, chosen per prompt: \emph{string-match} for free-form text answers, \emph{mathruler} for symbolic equivalence on numeric and algebraic answers, \emph{multiple-choice} for selected-letter answers, and \emph{gui-coordinate} for click-target predictions, where the reward decays smoothly with distance from the target.
The format score rewards a single \texttt{<think>} reasoning block followed by a single \texttt{\textbackslash boxed} answer, with partial credit when the policy emits extra reasoning or boxed entries.
This keeps correct answers from being zeroed out by the surface format errors common in VLM checkpoints after SFT, while still discouraging verbose multi-answer outputs.

To ensure an informative learning signal, we apply pass-rate filtering using 8 rollouts
per prompt from the initial policy checkpoint, retaining only prompts whose empirical
pass rate is below 0.8; prompts that are trivially solvable at initialization are discarded. The filtering is based on the same verifiers that are used during training.
We additionally include a small set of unanswerable or image-text-mismatched prompts to train the policy to abstain when visual evidence is insufficient.

The resulting corpus and verifier suite are inherited by OmniRL as the image component of its mixed-modality training mixture.

\subsubsection{Omni-RL}
Understanding is inherently challenging, and extending it to multiple modalities further increases complexity due to the need for sophisticated cross-modal reasoning. Prior advances in text reasoning have demonstrated the effectiveness of structured reasoning in improving model performance~\citep{wei2022chain,shao2024deepseekmath}. More recently, omni-modal reasoning has also been shown to be beneficial to omni and video tasks~\citep{ye2025omnivinci}. Motivated by these findings, we develop a unified reinforcement learning training stage aimed at enhancing the model’s capacity for coherent reasoning across images, videos, and audio modalities. 

To make omni RL training possible, we curate a diverse, omni-modal training corpus of approximately 120K prompts spanning 113 sub-datasets across four modality groups: image, video, audio, and text-only reasoning.
The dataset is constructed by aggregating and filtering data from multiple sources:
\textbf{(1)} \emph{Omni RL data} ($\sim$17.6K samples): synthetic data generated from video content with accompanying audio, covering diverse visual understanding and temporal reasoning tasks;
\textbf{(2)} \emph{Video RL data} ($\sim$8.5K): video-only question--answer pairs targeting spatial, temporal, and causal reasoning. 
\textbf{(3)} \emph{Image RL data} ($\sim$32K): a large-scale image understanding set drawing from OCR ($\sim$10.5K), chart analysis ($\sim$8.9K), game-related visual QA ($\sim$11.9K), GUI grounding ($\sim$7.1K), and additional curated domains;
\textbf{(4)} \emph{Audio RL data} ($\sim$4.2K) and \emph{ASR} ($\sim$3.8K): audio question-answering and automatic speech recognition tasks at various utterance lengths. 
We incorporate an ASR verifier to stabilize the speech recognition capability of our model. The reward is 1 - WER, where WER is computed after text normalization.

To ensure balanced difficulty and effective learning signals, we apply pass-rate filtering based on the initial policy checkpoint.
We retain only prompts on which the base model achieves a pass rate between 0.1 and 0.9 (with stricter 0.3--0.7 bands for AudioQA), thereby excluding prompts that are either trivially solvable or entirely intractable for the current policy.
The verification pipeline supports five task types: multiple-choice (34\%), string matching (31\%), mathematical rule-based verification (26\%), GUI coordinate grounding (6\%), and ASR evaluation (3\%).
We additionally include a small set of unanswerable or mismatched samples ($\sim$4K) to train the model to appropriately abstain when evidence is insufficient.

\subsubsection{RL Training Details}
Training is conducted on NVIDIA B200 and H100 GPU cluster using a Ray-based distributed training framework built on NeMo-RL \citep{nemo-rl}.
The global batch size is set to 4,096 with 16 rollouts for each prompt and a micro-batch size of 1.
 We apply an adapted version of Group Sequence Policy Optimization (GSPO)~\citep{zheng2025group,shao2024deepseekmath} as the RL training algorithm.

We use a multimodal deduplication strategy during the generation phase to leverage a unique multimodal tensor with rollouts associated with each prompt. We leverage tensor, expert, and context parallelism during training. All experiments are run with the AdamW optimizer with $\beta_1$ and $\beta_2$ set to $0.9$ and $0.999$, respectively, and a linear warmup.
\section{Experiments}
\label{section:experiments}

In Sections~\ref{sec:exp_vis}–\ref{sec:exp_text}, we conduct a comprehensive evaluation of the model’s ability to reason over vision, audio, and text inputs, and present the corresponding results. In Section~\ref{sec:exp_evs}, we analyze the efficiency gains achieved through Efficient Video Sampling (EVS) for video inputs. In Sections~\ref{sec:exp_quant} and \ref{sec:exp_inference}, we examine the impact of quantization on model accuracy and efficiency.

For the vision and audio evaluations in Sections~\ref{sec:exp_vis}–\ref{sec:exp_audio_vis}, we use the VLMEvalKit \citep{duan2025vlmevalkitopensourcetoolkitevaluating}\footnote{\url{https://github.com/open-compass/VLMEvalKit}} framework with a vLLM \citep{kwon2023efficient}\footnote{\url{https://github.com/vllm-project/vllm}} backend. Text evaluations are conducted using the NeMo-Skills\footnote{\url{https://github.com/NVIDIA-NeMo/Skills}} framework.

\subsection{Visual Evaluations}
\label{sec:exp_vis}

We conduct a comprehensive evaluation of our model on the following broad categories:
\begin{enumerate}
    \item \textbf{STEM Reasoning}: MMMU \citep{yue2023mmmu}, MathVista-Mini \citep{lu2024mathvista}
    \item \textbf{Document Understanding, OCR \& Charts}: MMLongBench-Doc \citep{ma2024mmlongbenchdocbenchmarkinglongcontextdocument}, OCRBench\citep{Liu_2024}, OCRBench-V2 \citep{fu2024ocrbenchv2improvedbenchmark}, ChartQA \citep{masry2022chartqabenchmarkquestionanswering}, AI2D \citep{kembhavi2016diagramworthdozenimages}, TextVQA \citep{singh2019towards}, DocVQA \citep{mathew2021docvqadatasetvqadocument}, InfoVQA \citep{mathew2021infographicvqa}, OCR-Reasoning \citep{huang2025ocrreasoningbenchmarkunveilingtrue}, CharXiv \citep{wang2024charxivchartinggapsrealistic}
    \item \textbf{Visual Grounding \& Spatial Reasoning}: TreeBench \citep{wang2025traceableevidenceenhancedvisual}, CV-Bench \citep{tong2024cambrian1fullyopenvisioncentric}, RefCOCO\citep{kazemzadeh2014referitgame}
    \item \textbf{GUI Understanding}: ScreenSpot \citep{cheng2024seeclickharnessingguigrounding}, ScreenSpot-v2 \citep{wu2024osatlasfoundationactionmodel}, ScreenSpot Pro \citep{li2025screenspotproguigroundingprofessional}, OSWorld \citep{OSWorld}
    \item \textbf{Video Understanding}: Video-MME \citep{fu2025videommefirstevercomprehensiveevaluation}
\end{enumerate}

As shown in Table~\ref{tab:vlm-benchmark-summary}, we observe significant improvements compared to \ourprevmodel across all benchmarks and even outperforms Qwen3-Omni on several categories.

\begin{table*}[htbp]
\centering
\resizebox{\textwidth}{!}{
\begin{tabular}{ll|cc|cc|c|c}
\toprule
\textbf{Task} & \textbf{Benchmark} & \multicolumn{2}{c|}{\textbf{\ourmodel}} & \multicolumn{2}{c|}{\textbf{\ourprevmodel}} & \textbf{Qwen3-Omni} & \textbf{Qwen3.5-Omni} \\
\midrule 
Open-Source & & \multicolumn{2}{c|}{\ding{51}} & \multicolumn{2}{c|}{\ding{51}} & \ding{51}& \ding{55} \\
Size & & \multicolumn{2}{c|}{30B-A3B} & \multicolumn{2}{c|}{12B} & 30B-A3B & Flash \\
Mode & & \makecell[c]{Reasoning\\off} & \makecell[c]{Reasoning\\on} & \makecell[c]{Reasoning\\off} & \makecell[c]{Reasoning\\on} & & \\
\midrule

\multirow{2}{*}{\makecell[l]{\textbf{STEM}\\\textbf{Reasoning}}} 
& MMMU (val) & 55.2 & 70.8 & 55.3 & 67.8 & 75.6 & \textbf{76.9} \\
& MathVista-Mini & 71.9 & 82.8 & 69.0 & 75.5 & 80.0 & \textbf{82.9}  \\
\midrule

\multirow{10}{*}{\makecell[l]{\textbf{Document}\\\textbf{Understanding,}\\\textbf{OCR \& Charts}}} 
& MMLongBench-Doc & 46.1 & \textbf{57.5} & 32.1 & 38.0 & 49.5 & 53.6 \\
& OCRBench &  88.3 & 86.6 & 85.6 & 83.5 & 86.0 & \textbf{89.1} \\
& OCRBenchV2 (EN/ZH) & 65.8/52.0 & \textbf{67.0}/\textbf{52.7} & 62.0/44.2 & 54.8/39.8 & - & -  \\
& ChartQA (Test) & 89.9 & \textbf{90.3} & 89.8 & 84.9 & 89.5 & -  \\
% & DocVQA (Val) & 93.9 & 95.5 & & & - & -  \\
& DocVQA (Test) & 93.3 & \textbf{95.6} & 94.7 & 93.2 & 95.3 & - \\
& AI2D (Test) & 88.5 & 88.5 & 87.2 & 84.7 & 86.62 & \textbf{89.0} \\
& TextVQA (Val) & 85.1 & 81.0 & \textbf{85.4} & 76.1 & 81.7 & -  \\
& InfoVQA (Test) & 83.6 & \textbf{86.8} & 79.4 & 80.4 & 83.31 &  - \\
% & InfoVQA (Val) &81.2 &85.7 & & & -  &  - \\
& OCR-Reasoning &22.2 & \textbf{54.14} & 21.0 & 33.9 & 49.9 & - \\
& CharXiv (RQ/DQ) & 49.1/81.9 & 63.6/\textbf{88.9} & 41.7/76.5 & 41.3/77.2 & 61.1/-  & \textbf{64.4}/- \\
\midrule

\multirow{2}{*}{\makecell[l]{\textbf{Visual Grounding \&}\\\textbf{Spatial Reasoning}}}
& TreeBench & 43.7& \textbf{51.6} & 38.5 & 42.5 & - & - \\
& CV-Bench & \textbf{84.2} & 84.0 & 81.0 & 78.3 & - & - \\
& RefCOCO & 80.6 & 90.5 & - & - & - & \textbf{92.6} \\
\midrule

\multirow{3}{*}{\textbf{GUI}} 
& ScreenSpot & \textbf{90.3} & 89.3 & 39.4 & 42.5 & - & -  \\
& ScreenSpot-v2 & \textbf{93.4} & 92.8 & 41.7 & 42.8 & - & -  \\
& ScreenSpot-Pro & 59.3 &57.8 & 4.8 & 5.5 & \textbf{59.7} & - \\
& OSWorld & - & \textbf{47.4} & - & 11.1 & 29.0 & - \\
\midrule

\makecell[l]{\textbf{Video}\\\textbf{Understanding}}
& VideoMME (w/o sub) &70.8 &72.2 & 66.0 & 63.0 & 70.5 & \textbf{77.0} \\

\bottomrule
\end{tabular}
}
\caption{Comparison of \ourmodel with our previous release, \ourprevmodel, as well as other state-of-the-art omni-modal models.
}
\label{tab:vlm-benchmark-summary}
\end{table*}

\subsection{Audio Evaluations}
\label{sec:exp_audio}

We evaluate our model across three broad categories:

\begin{enumerate}
    \item \textbf{Automatic Speech Recognition (ASR):} We use the OpenASR leaderboard~\citep{openasr}, and report word error rate on its English subset, including AMI, Earnings22, GigaSpeech, LibriSpeech, SPGISpeech, TED-LIUM and VoxPopuli. For long-form ASR, we additionally evaluate on TED-LIUM Longform~\citep{fox2024updated}, which tests transcription quality and long-context consistency on continuous speech.
    
    \item \textbf{Audio Understanding:} We evaluate on MMAU~\citep{mmau}, a benchmark of $\sim$10k audio clips with QA pairs spanning speech, environmental sounds, and music, covering 27 skills in information extraction and multi-step reasoning.
    
    \item \textbf{Voice Interaction \& Reasoning:} We use VoiceBench~\citep{voicebench}, which assesses LLM-based voice assistants on realistic spoken interactions, evaluating knowledge, instruction following, and safety across diverse speakers and environments.
\end{enumerate}

\begin{table*}[htb]
\centering
\scriptsize
\setlength{\tabcolsep}{3pt}
\renewcommand{\arraystretch}{0.9}
\begin{tabular}{lll|c|c|c}
\toprule
\textbf{Task} & \textbf{Benchmark} & \textbf{Subtask} & \textbf{\ourmodel} & \textbf{Qwen3-Omni} & \textbf{Qwen3.5-Omni} \\
\midrule
Open-Source & & & \ding{51} & \ding{51} & \ding{55} \\
Size &  &  & 30B-A3B & 30B-A3B & Flash \\
\midrule

\multirow{9}{*}{\textbf{ASR}}
& \multirow{9}{*}{\makecell[l]{OpenASR\\\footnotesize{(Reasoning off)}}}
& AMI & \textbf{11.09} & 12.52 & -- \\
& & Earnings22 & \textbf{11.27} & 12.3 & -- \\
& & GigaSpeech & 9.66 & \textbf{8.49} & -- \\
& & LibriSpeech (clean) & 1.57 & 1.52 & \textbf{1.3} \\
& & LibriSpeech (other) & 2.96 & 3.22 & \textbf{2.4} \\
& & SPGISpeech & \textbf{1.98} & 3.69 & -- \\
& & TED-LIUM & 3.44 & \textbf{2.38} & -- \\
& & VoxPopuli & \textbf{5.6} & 8.26 & -- \\
& & \textbf{OpenASR Avg} & \textbf{5.95} & 6.55 & -- \\
\midrule

\textbf{Long-form ASR}
& \makecell[l]{TED-LIUM\\\footnotesize{(Reasoning off)}}
& -- & 3.11 & \textbf{2.4} & -- \\
\midrule

\multirow{4}{*}{\textbf{Audio Understanding}}
& \multirow{4}{*}{\makecell[l]{MMAU\\\footnotesize{(Reasoning off)}}}
& Music & 74.2 & -- & -- \\
& & Audio & 76.9 & -- & -- \\
& & Speech & 72.8 & -- & -- \\
& & \textbf{MMAU Avg} & 74.6 & 77.5 & \textbf{80.4} \\
\midrule

\multirow{10}{*}{\textbf{Voice Interaction}}
& \multirow{10}{*}{\makecell[l]{VoiceBench\\\footnotesize{(Reasoning on)}}}
& IFEval & \textbf{88.7} & 80.6 & -- \\
& & BBH & \textbf{91.1} & 88.9 & -- \\
& & AdvBench & \textbf{100} & 97.2 & -- \\
& & AlpacaEval & 95.0 & \textbf{96.4} & -- \\
& & CommonEval & \textbf{91.3} & 90.5 & -- \\
& & WildVoice & \textbf{91.7} & 90.5 & -- \\
& & OpenBookQA & 93.0 & \textbf{94.3} & -- \\
& & MMSU & 82.3 & \textbf{83.0} & -- \\
& & SD-QA & 71.4 & \textbf{78.1} & -- \\
& & \textbf{VoiceBench Avg} & \textbf{89.4} & 88.8 & 87.8 \\

\bottomrule
\end{tabular}
\caption{Comparison of Nemotron 3 Nano Omni with other state-of-the-art open source models on diverse audio and speech tasks, ASR (OpenASR), long-form ASR (TED-LIUM), MMAU, and VoiceBench. ASR tasks are measured by word error rate (lower is better). For MMAU and VoiceBench, higher is better. ASR and MMAU use non-reasoning settings, while VoiceBench uses reasoning.}
\label{tab:audio-combined}
\end{table*}

As shown in Table~\ref{tab:audio-combined}, Nemotron 3 Nano Omni outperforms Qwen family models on ASR and VoiceBench benchmarks.

\subsection{Audio-Visual Evaluations}
\label{sec:exp_audio_vis}

We evaluate our model on audio-visual perception and reasoning using two complementary benchmarks:

\begin{enumerate}

 \item \textbf{DailyOmni}~\citep{zhou2025dailyomni}: an audio-visual QA benchmark for cross-modal reasoning in daily scenarios, with 684 videos (segmented into 30 and 60 secoond clips) and 1,197 multiple-choice questions across six tasks, testing temporal alignment, event understanding, causal reasoning, and cross-modal consistency.

 \item \textbf{WorldSense}~\citep{worldsense}: a large-scale omni-modal benchmark with 1,662 long-context videos and 3,172 multiple-choice questions across 26 tasks, evaluating long-range dependencies, sound grounding, temporal reasoning, and complex cross-modal inference.

\end{enumerate}

As shown in Table~\ref{tab:omni-benchmarks}, Nemotron 3 Nano Omni outperforms Qwen3-Omni on both reasoning on and off modes.

\begin{table*}[htbp]
\centering
\scriptsize
\setlength{\tabcolsep}{4pt}
\renewcommand{\arraystretch}{0.9}
\begin{tabular}{l|c c|c c|c}
\toprule
\textbf{Benchmark} 
& \multicolumn{2}{c|}{\textbf{\ourmodel}} 
& \multicolumn{2}{c|}{\textbf{Qwen3-Omni}} 
& \textbf{Qwen3.5-Omni} \\
\midrule

Open-Source 
& \multicolumn{2}{c|}{\ding{51}} 
& \multicolumn{2}{c|}{\ding{51}} 
& \ding{55} \\

Size 
& \multicolumn{2}{c|}{30B-A3B} 
& \multicolumn{2}{c|}{30B-A3B} 
& Flash \\

Mode
& Reasoning off & Reasoning on
& Instruct & Thinking
&  \\

\midrule

DailyOmni  
& 74.5 & 74.1 % on/off fill
&  71.9 & 73.6
& \textbf{81.8} \\

WorldSense 
& 55.2  & 55.4
& 54    & - 
&  \textbf{57.8} \\

\bottomrule
\end{tabular}
\caption{Comparison of Nemotron 3 Nano Omni with other state-of-the-art open source models on Video+Audio (Omni) benchmarks, measured by accuracy (higher is better).}
\label{tab:omni-benchmarks}
\end{table*}

\subsection{Text-only evaluations}
\label{sec:exp_text}

We conduct all pure-text evaluations with a maximum output length of 131,072 tokens, temperature set to 1.0, and top-p of 1.0. We report Pass@1 average of 8 runs for AIME-2025; an average of 4 runs for GPQA-Diamond \citep{rein2023gpqagraduatelevelgoogleproofqa}; and score of 1 run for SciCode \citep{tian2024scicoderesearchcodingbenchmark}, LiveCodeBench v5 (07/24 - 05/25) \citep{jain2024livecodebenchholisticcontaminationfree}, IFBench \citep{zhou2023instructionfollowingevaluationlargelanguage}, and TauBench V2 (telecom). We additionally include MMLU-Pro to assess general academic and knowledge-intensive reasoning.

Table \ref{tab:text-benchmarks} shows the evaluation on a selected text benchmarks compared to the \ourllm LLM that is used as the LLM backbone. The goal of the Omni model is maintain text benchmarks of the LLM while adding vision and audio understanding capabilities.

\begin{table*}[htbp]
\centering
\scriptsize
\setlength{\tabcolsep}{4pt}
\renewcommand{\arraystretch}{0.9}
\begin{tabular}{l|c|c|c}
\toprule
\textbf{Benchmark} & \textbf{\ourmodel} & \textbf{\ourllm} & \textbf{Qwen3-Omni} \\
\midrule
Open-Source & \ding{51} & \ding{51} & \ding{51} \\
Size & 30B-A3B & 30B-A3B & 30B-A3B \\
\midrule

MMLU-Pro              & 77.3 & \textbf{78.3} & 61.6 \\
GPQA (no tools)       & 72.2 & 73.0 & \textbf{73.1} \\
LiveCodeBench         & 63.2 & \textbf{68.3} & - \\
AIME25 (no tools)     & 82.1 & \textbf{89.1} & 73.7 \\
IFBench (prompt)      & \textbf{74.2} & 71.5 & - \\
AA-LCR                & \textbf{41.0} & 35.9 & - \\
TauBench V2 (Telecom) & \textbf{42.7} & 42.2 & - \\
SciCode               & 32.0 & \textbf{33.3} & - \\

\bottomrule
\end{tabular}
\caption{Comparison of Nemotron 3 Nano Omni, Nemotron 3 Nano LLM, and Qwen3-Omni across selected text-only benchmarks.}
\label{tab:text-benchmarks}
\end{table*}

\subsection{Reasoning budget control}
We study the effect of inference-time reasoning budgets by evaluating model performance under two settings: (1) a base configuration with a maximum sequence length of 16,384 tokens, and (2) a reasoning-enabled configuration with a 13K reasoning budget, a 1,024-token grace period, and a maximum sequence length of 16,384 tokens (Table~\ref{tab:reasoning-budget}).

Our results suggest that reasoning budget adjustment yields accuracy gains on select benchmarks under reasoning-on mode, with no degradation observed on the remaining ones. These gains with budget control may arise from the early termination of malformed reasoning traces with repetition loops on out-of-distribution tasks, as well as the truncation of overly verbose reasoning chains for problems requiring minimal or straightforward reasoning. 
\begin{table*}[htbp]
\centering
\scriptsize
\setlength{\tabcolsep}{2pt}
\renewcommand{\arraystretch}{0.9}
\begin{tabular}{l|c|c|c|c|c|c}   % <-- fixed: 7 columns
\toprule
\textbf{Benchmark} & \textbf{MathVista-Mini} & \textbf{MMLongBench-Doc} & \textbf{DocVQA (Val)} & \textbf{Charxiv(RQ)} &  \textbf{RefCOCO} & \textbf{VideoMME} \\
\midrule

w.o. reasoning budget & 80.3 & 54.5 & \textbf{95.3} & 61.8 & 90.4 & 67.5 \\
% w. reasoning budget   & 82.8 & 57.5  & 95.5  & 63.6  & 90.5  & 72.2 \\
w. reasoning budget   & \textbf{82.8} & \textbf{56.8} & 95.2  & \textbf{64} & \textbf{90.6}  & \textbf{70.3}  \\
\bottomrule
\end{tabular}
\caption{Effect of reasoning budget across several key benchmarks.}
\label{tab:reasoning-budget}
\end{table*}

\subsection{Conv3D and Efficient Video Sampling (EVS)}
\label{sec:exp_evs}

Nemotron 3 Nano Omni reduces the cost of long video inputs through two stacked mechanisms on the vision side. Conv3D is an architecture change applied during both training and inference: every $T=2$ consecutive frames are fused into a single ``tubelet'' before the first ViT block. This halves the number of vision tokens flowing through the ViT and the LLM, cutting both ViT prefill cost and LLM-side prefill, attention compute, and KV-cache footprint. EVS (Efficient Video Sampling)\citep{bagrov2025efficientvideosamplingpruning} is a runtime-only feature that drops video tokens after the ViT blocks and the vision adapter, immediately before they reach the LLM. For each spatial position $(h, w)$ it computes the cosine dissimilarity between consecutive tubelets and keeps the globally most-dissimilar tokens up to a budget set by the pruning rate $q$; the entire first tubelet is pinned to maximum dissimilarity so it is always retained as an anchor. The two mechanisms compose multiplicatively: Conv3D halves the number of tokens in the temporal dimension, EVS prunes the remaining tokens in the spatial dimension.

Table~\ref{tab:conv3d-evs-matrix} compares the four combinations of Conv3D
and EVS for both BF16 and NVFP4 checkpoints, with EVS fixed at $q=0.5$.
Each accuracy column reports per-benchmark scores at 128 and 256 sampled
frames, with reasoning off. Accuracy is averaged across three runs with identical settings. TTFT is averaged across five concurrency-1 \texttt{aiperf} runs
against a synthetic 512-frame, $512 \times 512$  video at 30~fps.

\begin{table*}[htbp]
\centering
\scriptsize
\setlength{\tabcolsep}{3pt}
\renewcommand{\arraystretch}{0.9}
\begin{tabular}{l|cc|cc|cc|cc|cc|c}
\toprule
\multirow{2}{*}{\textbf{Configuration}}
& \multicolumn{2}{c|}{\textbf{DailyOmni}}
& \multicolumn{2}{c|}{\textbf{LongVideoBench}}
& \multicolumn{2}{c|}{\textbf{Video-MME}}
& \multicolumn{2}{c|}{\textbf{WorldSense}}
& \multicolumn{2}{c|}{\textbf{Avg}}
& \textbf{TTFT} \\
& 128f & 256f & 128f & 256f & 128f & 256f & 128f & 256f & 128f & 256f & (ms) \\
\midrule
BF16                  & 74.74 & 74.77 & 66.23 & 67.90 & 69.13 & 69.70 & 54.80 & 54.50 & 66.23 & 66.72 & 7969 \\
BF16 + EVS            & 74.46 & 74.38 & 65.70 & 67.80 & 69.80 & 70.10 & 54.87 & 55.40 & 66.21 & 66.92 & 6452 \\
BF16 + Conv3D         & 74.41 & 74.24 & 66.30 & 67.20 & 68.70 & 70.70 & 54.83 & 54.43 & 66.06 & 66.64 & 5984 \\
BF16 + Conv3D + EVS   & 73.74 & 73.54 & 65.70 & 66.60 & 68.60 & 70.70 & 55.07 & 54.43 & 65.78 & 66.32 & 5313 \\
\midrule
NVFP4                 & 71.71 & 71.68 & 66.07 & 66.93 & 69.23 & 69.97 & 53.27 & 52.45 & 65.07 & 65.26 & 6885 \\
NVFP4 + EVS           & 71.65 & 71.76 & 65.50 & 67.30 & 69.80 & 70.93 & 52.90 & 52.60 & 64.96 & 65.65 & 5977 \\
NVFP4 + Conv3D        & 70.37 & 70.84 & 65.90 & 66.43 & 68.70 & 70.30 & 52.63 & 52.27 & 64.40 & 64.96 & 5635 \\
NVFP4 + Conv3D + EVS  & 70.76 & 70.65 & 64.97 & 66.50 & 68.47 & 70.17 & 52.50 & 52.70 & 64.17 & 65.00 & 5083 \\
\bottomrule
\end{tabular}
\caption{Per-benchmark accuracy (128 frames / 256 frames) and TTFT at
concurrency 1 across Conv3D and EVS combinations, with EVS rate $q=0.5$, reasoning off.}
\label{tab:conv3d-evs-matrix}
\end{table*}

Both mechanisms significantly reduce TTFT on BF16: Conv3D
alone drops it from 7969~ms to 5984~ms ($-25$\%), EVS alone drops it to
6452~ms ($-19$\%), and stacking them yields 5313~ms ($-33$\% versus the
baseline) at a cost of about half a point of average accuracy. The same
ordering holds on NVFP4. Underlying these gains is a substantial reduction
in the number of input tokens for the LLM: a 512-frame video produces $\sim$141k input tokens
without either mechanism, drops to $\sim$75k with Conv3D enabled
($-47$\%), and drops further to $\sim$42k with Conv3D combined with EVS
at $q=0.5$ ($-70$\% versus the baseline).

Table~\ref{tab:evs-sweep} sweeps the EVS pruning rate $q$ on BF16 with
Conv3D enabled. Per-benchmark accuracy is essentially flat through $q=0.7$,
slightly reduces at $q=0.8$, and drops noticeably beyond, with
LongVideoBench being the most sensitive benchmark to aggressive pruning. TTFT
improves monotonically through the range, for a $\sim$$14$\% reduction at
$q=0.7$ versus no-EVS.

\begin{table*}[htbp]
\centering
\scriptsize
\setlength{\tabcolsep}{3pt}
\renewcommand{\arraystretch}{0.9}
\begin{tabular}{l|cc|cc|cc|cc|cc|c}
\toprule
\textbf{EVS}
& \multicolumn{2}{c|}{\textbf{DailyOmni}}
& \multicolumn{2}{c|}{\textbf{LongVideoBench}}
& \multicolumn{2}{c|}{\textbf{Video-MME}}
& \multicolumn{2}{c|}{\textbf{WorldSense}}
& \multicolumn{2}{c|}{\textbf{Avg}}
& \textbf{TTFT} \\
\multicolumn{1}{c|}{$q$} & 128f & 256f & 128f & 256f & 128f & 256f & 128f & 256f & 128f & 256f & (ms) \\
\midrule
none  & 74.41 & 74.24 & 66.30 & 67.20 & 68.70 & 70.70 & 54.83 & 54.43 & 66.06 & 66.64 & 5984 \\
0.5   & 73.74 & 73.54 & 65.70 & 66.60 & 68.60 & 70.70 & 55.07 & 54.43 & 65.78 & 66.32 & 5313 \\
0.6   & 74.41 & 74.44 & 65.10 & 66.50 & 68.90 & 70.90 & 54.57 & 54.40 & 65.74 & 66.56 & 5173 \\
0.7   & 73.82 & 73.77 & 65.00 & 65.40 & 68.70 & 70.30 & 54.10 & 53.80 & 65.41 & 65.82 & 5124 \\
0.8   & 73.38 & 73.24 & 64.30 & 64.80 & 67.80 & 70.10 & 54.00 & 53.73 & 64.87 & 65.47 & 5182 \\
0.9   & 71.54 & 71.71 & 59.80 & 61.00 & 67.10 & 68.30 & 52.97 & 52.87 & 62.85 & 63.47 & 4883 \\
0.95  & 69.31 & 69.31 & 55.60 & 57.90 & 64.60 & 66.30 & 51.67 & 51.93 & 60.29 & 61.36 & 4804 \\
\bottomrule
\end{tabular}
\caption{BF16 with Conv3D enabled, varying EVS pruning rate $q$, reasoning off. Same
column structure as Table~\ref{tab:conv3d-evs-matrix}.}
\label{tab:evs-sweep}
\end{table*}

\subsection{Quantization}
\label{sec:exp_quant}

Inspired by the quantization recipe from Nemotron 3 Super, we pursued a
mixed-precision strategy for FP4: routed MoE experts are quantized to NVFP4
(FP4 E2M1 values with per-block FP8 E4M3 scales over groups of 16 elements and
an additional per-tensor FP32 global scale), while the Mamba \texttt{in\_proj}
/ \texttt{out\_proj}, shared experts, and attention \texttt{o\_proj} are
quantized to FP8 (per-tensor E4M3 values with a per-tensor FP32 scale). All
remaining language-model layers are left in BF16, as are the vision and audio
encoders and their MLP projectors. For the KV cache we use FP8, while the Mamba
SSM state cache is kept at FP32 at serving time. This gives a model-weight
footprint of 4.98 effective bits per weight (20.9 GB vs the 61.5 GB BF16
reference). For FP8 we quantize every linear layer in the language model to
per-tensor E4M3 (with a per-tensor FP32 scale), with the exception of the MoE
router and \texttt{lm\_head}, and pair it with an FP8 KV cache. The vision and
audio encoders and their MLP projectors are excluded entirely. This yields
$\sim$8.5 bpw (32.8 GB). We evaluated the quantized models across 25 text,
image, video, and audio benchmarks and found a median accuracy drop of less
than $1\%$ vs BF16 for both FP8 and NVFP4.

\begin{table*}[htbp]
\centering
\scriptsize
\setlength{\tabcolsep}{4pt}
\renewcommand{\arraystretch}{0.9}
\begin{tabular}{l|c|c|c}
\toprule
\textbf{Benchmark} & \textbf{BF16} & \textbf{FP8} & \textbf{NVFP4} \\
\midrule
Size (GB)       & 61.5  & 32.8  & 20.9 \\
Effective bpw   & 16.00 & 8.5 & 4.98 \\
\midrule
MathVista-Mini                       & 71.90 & 71.05 & 71.30 \\
CharXiv (RQ)                         & 49.10 & 48.05 & 47.95 \\
OCR-Reasoning                        & 22.20 & 23.43 & 22.78 \\
MMLongBench-Doc                      & 46.10 & 45.84 & 45.78 \\
OCRBenchV2 (EN)                      & 65.80 & 65.63 & 65.77 \\
OCRBenchV2 (ZH)                      & 52.00 & 50.24 & 50.39 \\
CV-Bench                             & 84.20 & 85.62 & 85.27 \\
VideoMME                             & 70.80 & 69.40 & 69.60 \\
DailyOmni                            & 74.50 & 74.06 & 74.23 \\
WorldSense-AVLM                      & 55.20 & 54.40 & 54.60 \\
MMAU                                 & 74.62 & 74.56 & 74.34 \\
\midrule
TedLium-Longform (WER$\downarrow$)           & 3.11 & 3.12 & 3.04 \\
HF-ASR avg, 8 short-form (WER$\downarrow$)   & 5.95 & 5.97 & 5.95 \\
\midrule
\textbf{Mean (11 non-ASR)}       & \textbf{60.58} & \textbf{60.21} & \textbf{60.18} \\
\textbf{Median (11 non-ASR)}     & \textbf{65.80} & \textbf{65.63} & \textbf{65.77} \\
\textbf{$\Delta$ vs BF16 (mean)} & ---   & $-0.37$ & $-0.40$ \\
\bottomrule
\end{tabular}
\caption{Accuracy of Nemotron 3 Nano Omni at BF16, FP8, and NVFP4
across the eval suite.}
\label{tab:quant-accuracy}
\end{table*}

%\ehsan{voicebench is missing in this table}

\subsection{Inference efficiency}
\label{sec:exp_inference}

\noindent
\textbf{NVFP4. }
Compared to BF16 precision, NVFP4 on NVIDIA B200 provides up to 7.5$\times$ the output token throughput at iso-interactivity (18200 tok/s versus 2400 tok/s, at 150 tok/s/user) on an single-image reasoning usecase.

\noindent
\textbf{Low-latency single-stream inference. }
\ourmodel delivers strong single-stream inference performance on NVIDIA B200, reaching more than
500 output tokens/s at a concurrency of 1. This low-latency generation rate is sustained at longer sequence lengths and
with larger multimodal inputs, such as long videos or multi-document workloads, due to the hybrid
architecture. This is approximately 2.4--2.9$\times$ faster than Qwen3-Omni, which reaches
175--210 output tokens/s depending on input size and sequence length, and 2$\times$ faster than
\ourprevmodel, which reaches 250 output tokens/s.

For a multi-document workload, \ourmodel achieves a time-to-first-token (TTFT) of
approximately 1.3 s, compared to more than 2.5 s for Qwen3-Omni.

\noindent
\textbf{High-throughput serving.}
At maximum concurrency on a single NVIDIA B200, \ourmodel reaches 5000 output tokens/s on a
multi-document workload. At an iso-interactivity target of 50 output tokens/s per user, the
deployment provides 9$\times$ higher output throughput than Qwen3-Omni on long-video workloads and
7.5$\times$ higher output throughput on multi-document workloads. Compared with \ourprevmodel,
\ourmodel provides 3$\times$ higher throughput at the same interactivity target.

\noindent
\textbf{Experimental setup.}
All measurements use a single NVIDIA B200 GPU and vLLM nightly as of 2026-04-19 with EVS 50\%.
\ourmodel is evaluated in NVFP4, Qwen3-Omni with dynamic FP8 quantization, and \ourprevmodel in
NVFP4. Text ISL=50 and OSL=8000. The multi-document workload contains 32
images at 1024$\times$1536 resolution. The long-video workload contains 512 frames at
512$\times$512 resolution.
\section{Conclusion}
\label{section:conclusion}

We introduced \ourmodel, an efficient omni-modal model that extends the Nemotron multimodal family with native audio support and consistently stronger reasoning across text, images, video, and audio. Built on the \ourllm MoE hybrid backbone and augmented with the C-RADIOv4-H vision encoder and the Parakeet-TDT audio encoder, the model combines dynamic image resolution, Conv3D-based temporal video compression, and a 256K context length to process long, heterogeneous multimodal inputs with high accuracy. We use a multi-stage training recipe that progressively introduces new modalities and extends context to enable robust cross-modal alignment while preserving the text reasoning ability of the base LLM.
 
Across a broad evaluation suite, \ourmodel delivers consistent gains over \ourprevmodel and achieves leading or competitive results on document understanding (OCRBench-V2, MMLongBench-Doc, ChartQA, CharXiv), agentic GUI use (ScreenSpot, ScreenSpot-Pro, OSWorld), long audio-video comprehension (WorldSense, DailyOmni), and voice interaction (VoiceBench), while retaining the text reasoning performance of the \ourllm backbone. Combined with innovative multimodal token-reduction techniques, these capabilities translate into substantially lower inference latency and several-fold higher throughput than comparably sized models. We release model checkpoints in BF16, FP8, and FP4 formats alongside a large portion of our training data and code, with the goal of enabling the community to further advance efficient omni-modal modeling.
\section{Contributors}

\textbf{Core Model Development}

Amala Sanjay Deshmukh, Kateryna Chumachenko, Tuomas Rintamaki, Matthieu Le, Tyler Poon, Danial Mohseni Taheri, Ilia Karmanov, Guilin Liu, Jarno Seppanen, Arushi Goel, Mike Ranzinger, Greg Heinrich, Guo Chen, Lukas Voegtle, Philipp Fischer, Timo Roman, Karan Sapra, Collin McCarthy, Shaokun Zhang, Fuxiao Liu, Hanrong Ye, Yi Dong, Mingjie Liu, Yifan Peng, Piotr Zelasko, Zhehuai Chen, Nithin Rao Koluguri, Nune Tadevosyan, Lilit Grigoryan, Ehsan Hosseini Asl, Pritam Biswas, Leili Tavabi, Yuanhang Su, Zhiding Yu, Peter Jin, Alexandre Milesi, Netanel Haber

\textbf{Data Generation and Curation}

Yao Xu, Sarah Amiraslani, Nabin Mulepati, Eric Tramel, Jaehun Jung, Ximing Lu, Brandon Cui, Jin Xu, Zhiqi Li, Shihao Wang, Yuanguo Kuang, Shaokun Zhang, Huck Yang, Boyi Li, Hongxu (Danny) Yin, Song Han, Bilal Kartal, Pavlo Molchanov, Adi Renduchintala, Charles Wang,, David Mosallanezhad, Soumye Singhal, Luis Vega, Katherine Cheung, Sreyan Ghosh, Yian Zhang, Alexander Bukharin, Venkat Srinivasan, Johnny Greco, Andre Manoel, Maarten Van Segbroeck, Suseella Panguliri, Rohit Watve, Divyanshu Kakwani, Shubham Pachori, Jeffrey Glick, Radha Sri-Tharan, Aileen Zaman, Khanh Nguyen, Shi Chen, Jiaheng Fang, Qing Miao, Wenfei Zhou, Yu Wang, Zaid Pervaiz Bhat, Varun Praveen, Arihant Jain, Ramanathan Arunachalam, Tomasz Kornuta, Ashton Sharabiani, Amy Shen, Wei Huang

\textbf{Systems, Data and Infrastructure}

Yi-Fu Wu, Ali Roshan Ghias, Huiying Li, Brian Yu, Nima Tajbakhsh, Chen Cui, Wenwen Gao, Li Ding, Terry Kong, Manoj Kilaru, Anahita Bhiwandiwalla, Marek Wawrzos, Daniel Korzekwa, Pablo Ribalta, Grzegorz Chlebus, Besmira Nushi, Ewa Dobrowolska, Maciej Jakub Mikulski, Kunal Dhawan, Steve Huang, Jagadeesh Balam, Yongqiang Wang, Nikolay Karpov, Valentin Mendelev, George Zelenfroynd, Meline Mkrtchyan, Qing Miao, Omri Almog, Bhavesh Pawar, Rameshwar Shivbhakta, Sudeep Sabnis, Ashrton Sharabiani, Negar Habibi, Geethapriya Venkataramani, Pamela Peng, Prerit Rodney, Serge Panev, Richard Mazzarese, Nicky Liu, Michael Fukuyama, Andrii Skliar, Roger Waleffe, Duncan Riach, Yunheng Zou, Jian Hu, Hao Zhang, Binfeng Xu, Yuhao Yang, Zuhair Ahmed

\textbf{Inference and Optimization}

Alexandre Milesi, Carlo del Mundo, Chad Voegele, Zhiyu Cheng, Nave Assaf, Andrii Skliar, Daniel Afrimi, Natan Bagrov, Ran Zilberstein, Ofri Masad, Eugene Khvedchenia, Natan Bagrov, Borys Tymchenko, Tomer Asida, Daniel Afrimi, Parth Mannan, Victor Cui

\textbf{Safety}

Michael Evans, Katherine Luna, Jie Lou, Pinky Xu, Guyue Huang, Negar Habibi, Michael Boone, Pradeep Thalasta, Adeola Adesoba, Dina Yared, Christopher Parisien, Leon Derczynski, Shaona Ghosh, Wes Feely, Micah Schaffer, Radha Sri-Tharan, Jeffrey Glick, Barnaby Simkin, George Zelenfroynd, Tomasz Grzegorzek, Rishabh Garg

\textbf{Evaluation, Product and Legal}

Aastha Jhunjhunwala, Sergei Kolchenko, Farzan Memarian, Haran Kumar, Shiv Kumar, Isabel Hulseman, Anjali Shah, Kari Briski, Padmavathy Subramanian, Joey Conway, Udi Karpas, Jane Polak Scowcroft, Annie Surla, Shilpa Ammireddy, Ellie Evans, Jesse Oliver, Tom Balough, Chia-Chih Chen, Sandip Bhaskar, Alejandra Rico, Bardiya Sadeghi, Seph Mard, Katherine Cheung, Meredith Price, Laya Sleiman, Saori Kaji, Wesley Helmholz, Wendy Quan

\textbf{Leadership}

Michael Lightstone, Jonathan Cohen, Jian Zhang, Oleksii Kuchaiev, Boris Ginsburg, Jan Kautz, Eileen Long, Mohammad Shoeybi, Mostofa Patwary, Oluwatobi Olabiyi, Andrew Tao, Bryan Catanzaro, Udi Karpas

\bibliography{references}
\bibliographystyle{references}

\end{document}